\documentclass[12pt,letterpaper]{article}
\usepackage[utf8]{inputenc}
\usepackage{amsmath}
\usepackage{mathptmx, courier,textcomp}
\usepackage{amsfonts}
\usepackage{amssymb}
\usepackage{graphicx}
\usepackage{subcaption}
\usepackage{cite}
\usepackage{booktabs}
\usepackage{ulem}
\usepackage{tablefootnote}
\usepackage{boldline}
\usepackage{caption}
\usepackage{slashbox}
\usepackage{epstopdf}
\usepackage[top=1in,bottom=1in,left=1in,right=1in]{geometry} 
\usepackage{float}
\usepackage{hyperref}
\usepackage{booktabs}
\usepackage[export]{adjustbox}
\usepackage{float}
\usepackage{multirow}
\usepackage{multicol}
\usepackage{pifont}
\usepackage[ ]{authblk}
\usepackage{chemarrow}
\usepackage{comment}
\usepackage{makecell}
\usepackage{orcidlink}
\UseRawInputEncoding
\hypersetup{
    colorlinks=true,
    linkcolor=blue,
    filecolor=magenta,      
    urlcolor=blue,
    pdftitle={Overleaf Example},
    pdfpagemode=FullScreen,
}
\usepackage{multicol}
\begin{document}
%
% --------------------------------------------- Start Title
%
\title{CAE-Net: Generalized Deepfake Image Detection using Convolution and Attention Mechanisms with Spatial and Frequency Domain Features}
%
% --------------------------------------------- End Title
% 

%
% ---------------------------------------------  Start authors
%
\author[1]{Anindya Bhattacharjee\orcidlink{0009-0006-6496-7896}}
\author[1]{Kaidul Islam\orcidlink{0009-0002-2517-486X}}
\author[1]{Kafi Anan\orcidlink{0009-0005-9993-6523}}
\author[1]{Ashir Intesher\orcidlink{0009-0007-3372-6755}}
\author[1]{Abrar Assaeem Fuad\orcidlink{0009-0004-5114-428X}}
\author[1,2]{Utsab Saha\orcidlink{0000-0003-2106-8648}}
\author[1,*]{Hafiz Imtiaz\orcidlink{0000-0002-2042-5941}}

\affil[1]{\small{Department of Electrical and Electronic Engineering, BUET, 
Dhaka 1205, Bangladesh}}
\affil[2]{\small{Department of Computer Science and Engineering\\
BRAC University\\
Dhaka 1212, Bangladesh}}

\affil[*]{\small{\it{hafizimtiaz@eee.buet.ac.bd}}}

%
% ---------------------------------------------  end authors 
%

% ---------------------------------------------
\date{}
\maketitle
\begingroup
\renewcommand\thefootnote{}\footnote{This manuscript has been published in the \textit{Journal of Visual Communication and Image Representation} by Academic Press.}\addtocounter{footnote}{-1}
\endgroup

% -----------------------------------------
% --------------------------------------------
\sloppy
%
% --------------------------------------------- Start abstract
%
\begin{abstract}
The spread of deepfakes poses significant security concerns, demanding reliable detection methods. However, diverse generation techniques and class imbalance in datasets create challenges. We propose CAE-Net, a Convolution- and Attention-based weighted Ensemble network combining spatial and frequency-domain features for effective deepfake detection. The architecture integrates EfficientNet, Data-Efficient Image Transformer (DeiT), and ConvNeXt with wavelet features to learn complementary representations. We evaluated CAE-Net on the diverse IEEE Signal Processing Cup 2025 (DF-Wild Cup) dataset, which has a 5:1 fake-to-real class imbalance. To address this, we introduce a multistage disjoint-subset training strategy, sequentially training the model on non-overlapping subsets of the fake class while retaining knowledge across stages. Our approach achieved $94.46\%$ accuracy and a $97.60\%$ AUC, outperforming conventional class-balancing methods. Visualizations confirm the network focuses on meaningful facial regions, and our ensemble design demonstrates robustness against adversarial attacks, positioning CAE-Net as a dependable and generalized deepfake detection framework.
\end{abstract}


%
% --------------------------------------------- End abstract
%
% \pagebreak
%
%%%%%%%%%%%%%%%%%%%%%%%%%%%%% body  %%%%%%%%%%%%%%%%%%%%%%%%%%%%%%%%
%

%
% --------------------------------------------- Start introduction
%
\section{Introduction}
Deepfakes are media that are edited or created by generative Artificial Intelligence (AI), which do not exist or have not occurred in reality. Although deepfakes have existed for a long time, it took experts, and often entire studios, to generate \emph{convincing} fake media. Due to the surge of machine learning and generative AI tools, the generation of deepfakes has never been easier and more accessible. Such ease of access has raised serious societal concerns since these are being used to manipulate public opinion, commit fraud, create non-consensual pornography, and attack individuals through defamation or identity threats~\cite{doi:10.1089/cyber.2020.0272}. Deepfakes are also being used to gain malicious political advantage worldwide~\cite{westerlund2019emergence}. Although deepfake generation tools are widely accessible, detection tools are not nearly as effective and accurate. Thus, effective detection of deepfakes is a task of great importance and has been extensively studied for the past decade.

The combination of multiple deep learning architectures has been shown to perform well on deepfake detection. Gowda and Thillaiarasu~\cite{gowda2022investigation} showed that ensemble techniques of Convolutional Neural Network (CNN) models (ResNeXt~\cite{Xie_2017_CVPR} and Xception~\cite{chollet2017xception}) achieve higher accuracy on deepfake detection. Wodajo and Atnafu~\cite{wodajo2021deepfake} combined VGG-16~\cite{simonyan2014very} for feature extraction and Vision Transformer (ViT)~\cite{sharir2021imageworth16x16words} for classification, showing promising results on the DFDC dataset. Wolter et al.~\cite{wolter2022wavelet} proposed a wavelet-packet-based analysis for detecting Generative Adversarial Networks (GAN)-generated images, highlighting spatial frequency differences in synthetic content and achieving competitive results using combined architectures of wavelets and CNNs. Ricker et al.~\cite{ricker2022towards} proposed the usage of frequency domain features extracted via Discrete Cosine Transform (DCT)~\cite{ahmed2006discrete}, and achieved high accuracy rates (97.7\% for GAN and 73\% for diffusion models), highlighting the utility of frequency-domain artifacts. Patel et al. developed an enhanced deep CNN (D-CNN) architecture for deepfake detection, achieving reasonable accuracy and high generalization capability~\cite{patel2023improved}. Patch-Forensics, a technique analyzing smaller patches to detect local image artifacts, achieves 100\% average precision on StyleGAN-generated images using an Xception Block 2 classifier~\cite{frank2020leveraging}. However, Abdullah et al.~\cite{abdullah2024analysis} highlighted two major limitations of the existing networks, arguing that the state-of-the-art models often lack control over content and image quality, making them ineffective at detecting high-quality deepfakes if trained on low-quality fake images, and they also lack adversarial evaluation, where an attacker can exploit knowledge of the defense. Works on adversarial perturbation attacks to fool deep neural network (DNN) based detectors are a challenge in deepfake detection~\cite{hussain2021adversarial, gandhi2020adversarial}. Training on diverse and adversarially perturbed datasets can increase the robustness of a detection architecture~\cite{nguyen2022deep}. Lin et al. proposed a CNN-based method for deepfake detection that combines multi-scale convolution with a vision transformer (ViT)~\cite{lin2023deepfake}. Their approach features a multi-scale module incorporating dilation convolution and depth-wise separable convolution. This design aims to capture more facial details and identify tampering artifacts at various scales~\cite{lin2023deepfake}. Note that the ViT has gained significant popularity in recent years for deepfake image detection~\cite{heo2023deepfake, usmani2024efficient}. 

Researchers are not only relying on complex network architectures but also exploring other approaches, such as the Frequency Enhanced Self-Blending Images (FSBI) method for deepfake detection, which utilizes the Discrete Wavelet Transform (DWT)~\cite{hasanaath2025fsbi}. Zhu et al. introduced a learnable Image Decomposition Module (IDM) that uses a recomposition operation to highlight illumination inconsistencies~\cite{zhu2024deepfake}. It features a multi-level enhancement technique for effective feature recomposition and is trained in the logarithmic domain, with an extensible design~\cite{zhu2024deepfake}. To address challenges with low-quality datasets and inadequate detection performance, Cheng et al. have developed MSIDSnet~\cite{cheng2024deepfake}. This framework utilizes a multi-scale fusion (MSF) module to capture forged facial features and an interactive dual-stream (IDS) module to enhance feature integration across frequency and spatial domains~\cite{cheng2024deepfake}. Another recent detection model by Byeon et al. uses a spatial-frequency joint dual-stream convolutional neural network with learnable frequency-domain filtering kernels~\cite{byeon2024deep}.\\

\noindent\textbf{Research gap and our contribution. }We note that the existing detection methods only work well when the test image resembles the features of the images in the training dataset of the detector. As such, the major issues in the existing detection approaches are i) the lack of generalizability of extracted features, and ii) the inability to be up to date with new fake image generation techniques~\cite{kaur2024deepfake}. Motivated by this limitation, we propose a weighted ensemble-based deepfake detection framework, \emph{CAE-Net}, which combines convolutional and attention mechanisms to capture both spatial and frequency-domain cues. The model integrates three different neural network architectures with complementary strengths: i) an EfficientNet architecture~\cite{tan2019efficientnet}, ii) a Data-Efficient Image Transformer (DeiT)~\cite{touvron2021training}, and iii) a ConvNeXt model~\cite{liu2022convnet} enhanced with a wavelet-based preprocessing stage. This hybrid configuration, to the best of our knowledge, is the first to incorporate wavelet-domain feature enhancement within a CNN–Transformer ensemble, effectively leveraging local, global, and frequency-domain representations for generalized deepfake image detection. Since training on highly diverse datasets remains difficult for most existing architectures, we employ the IEEE Signal Processing Cup 2025 (DF-Wild Cup) dataset to demonstrate our proposed CAE-Net’s generalization capabilities. This dataset combines eight publicly available deepfake benchmark datasets under the DeepfakeBench framework and exhibits a pronounced class imbalance (5:1 fake-to-real ratio). To address this challenge, we design a disjoint set-based multistage training strategy, where the fake class is divided into five non-overlapping subsets and trained sequentially along with real samples, using weight warm-starting across stages to retain learned knowledge. The final weighted ensemble employs probability-level late fusion to combine the outputs of the three backbones, leading to statistically significant improvements over individual models and conventional ensemble techniques. As we demonstrate in Section~\ref{sec:result}, the proposed model offers strong classification performance and maintains robustness under adversarial perturbations, confirming the effectiveness and reliability of CAE-Net in generalized deepfake image detection.

The rest of the paper is organized as follows: Section \ref{sec:method} contains data preprocessing techniques, a description of our CAE-Net architecture, consisting of EfficientNet, DeiT, and ConvNeXt. Section \ref{sec:result} contains the classification results of multiple standard models on the dataset, an ablation study on the different ensemble variations, along with the implementation details and other evaluation metrics of our proposed approach. This also contains a comparison of our model's generalization capabilities against multiple existing works and a defense against adversarial perturbations. Section ~\ref{sec: Discussion} provides an overall view of the interpretation of the results, the significance of the findings, and the limitations of our study. Finally, Section \ref{sec:conc} draws concluding remarks on the contributions presented in the paper.

\section{Proposed Methodology: Weighted Ensemble of Neural Network Models}
\label{sec:method}
\subsection{Description of the Dataset}
As mentioned before, the \textit{DFWild-Cup} competition has provided a diverse dataset. The dataset contains images from eight publicly available standard datasets designed for the DeepfakeBench evaluation~\cite{DeepfakeBench_YAN_NEURIPS2023}: Celeb-DF-v1~\cite{li2020celeb}, Celeb-DF-v2~\cite{li2020celeb}, FaceForensics++~\cite{rossler2019faceforensics++}, DeepfakeDetection~\cite{Dufour_Gully_2019}, FaceShifter~\cite{li2019faceshifter}, UADFV~\cite{li2018ictu}, Deepfake Detection Challenge Preview~\cite{dolhansky2019dee}, and Deepfake Detection Challenge~\cite{dolhansky2020deepfake}. To ensure that the detector performance is not biased by specific pre-processing techniques, all datasets underwent uniform pre-processing. Additionally, any information that would identify the origin of the dataset has been eliminated, and the file names have been anonymized. For training, there are a total of 42,690 real images and 219,470 fake images. The validation dataset consists of 1,548 real images and 1,524 fake images. It is evident that there is a significant class imbalance present in the training data, as the ratio of fake and real images is approximately 5:1.

\subsection{Data Preprocessing}
\subsubsection{Class Balancing for Model Training}
\label{class balancing}
One of the key contributions of our paper is to devise a disjoint set-based multistage training approach to address the class imbalance present in the given dataset. Traditional class-balancing techniques (oversampling, SMOTE, class weighting) present significant limitations for our dataset's extreme imbalance.

Random oversampling: We empirically validated that replicating 42,690 real images to match 219,470 fakes introduce severe overfitting during model training, as models see identical real samples multiple times across epochs, memorizing specific artifacts rather than learning generalizable features.

Synthetic generation (SMOTE, GANs): Generating $\sim$177,000 synthetic real images risks introducing unrealistic artifacts or distribution shift. Moreover, since the task itself is deepfake detection, authentic real images must represent unmanipulated content. Synthetic augmentation could paradoxically teach models to accept fake-like patterns as real.

Class weighting: While computationally efficient, loss reweighting alone does not prevent the model from being overwhelmed by the sheer volume of fake samples during training, leading to bias toward the majority class despite adjusted gradients.

Our disjoint multistage approach addresses these issues by ensuring a balanced exposure where each stage trains on a 1:1 balanced subset (42,690 real + 43,894 fake images). All real images remain genuine and unmanipulated samples, preserving the true distribution of authentic content. This strategy effectively leverages our full dataset (262,160 images) without the artifacts, overfitting, or distribution shifts inherent to conventional balancing techniques applied to such extreme imbalances.

Since the number of fake images is approximately five times the number of real images, we divided the fake images $\{\mathcal{F}\}$ into five disjoint subsets $\{\mathcal{F}_1, \ldots, \mathcal{F}_5\}$. Each of the three aforementioned models, denoted as \( M_1, M_2, M_3 \), is trained in five stages with these disjoint subsets of fake images. At each stage \( i \), the model is trained using the entire real image set \( \mathcal{R} \) and one subset of fake images \( \mathcal{F}_i \), such that the training dataset of stage $i$ is given by:
\begin{equation}
    \mathcal{D}_i = \mathcal{R} \cup \mathcal{F}_i, \quad i \in \{1,2,3,4,5\}.
\end{equation}
We emphasize that the same set of real images $\{\mathcal{R}\}$ is treated as one class during each stage. The training of each model is initialized with pre-trained weights \( \theta_0 \), obtained from an ImageNet~\cite{deng2009imagenet}-trained version of the corresponding architecture. The model parameters at each stage are updated iteratively using the dataset \( \mathcal{D}_i \), through the optimization:
\begin{equation}
    \theta_i = \arg\min_{\theta} \sum_{(x,y) \in \mathcal{D}_i} \mathcal{L}(M(x; \theta), y),
\end{equation}
where \( \mathcal{L} \) represents the chosen loss function. After training on \( \mathcal{D}_i \), the final weights \( \theta_i \) from each stage are used to warm-start the next stage:
\begin{equation}
    \theta_{i+1} \leftarrow \theta_i, \quad \text{for } i \in \{1,2,3,4\}.
\end{equation}
This process continues until the model has been trained on all subsets of fake images, ensuring that the model learns from diverse variations of the fake images while preserving knowledge of real images. While training our models across stages, we maintained the hyperparameter settings stated in Table~\ref{tab:Hyperparameters}.

\subsubsection{Data Augmentation and Normalization}
As mentioned before, we utilized the same set of real images across five training stages to address the class imbalance. However, this evidently increases the chance of overfitting. To mitigate overfitting and enhance the model's generalization performance, we implemented four data augmentation techniques: random horizontal flips(with 50\% probability), random rotations(within $0^o-10^o$), color jittering (20\% variation in brightness, contrast, saturation and hue), and random resized cropping where it produced $224\times 224$ images, while randomly scaling the crop area between 80\% and 120\% of the original image size. These augmentations enhanced the diversity of the training dataset, reduced overfitting, and improved the model’s ability to generalize across a wider range of deepfake detection scenarios. It should be noted that such augmentations were used only in training the DeiT architecture. For EfficientNet, data augmentation didn’t enhance validation accuracy or generalization, likely due to its convolutional design already effectively captures the local features. In contrast, DeiT showed notable gains in validation accuracy with augmentation, as its attention-based architecture thrives on greater data variety to learn global patterns~\cite{xue2024differentiable}. Therefore, we applied augmentation only to DeiT in the final model. For the ConvNeXt, we employed wavelet transform-based feature extraction from the raw images, and data augmentation techniques would result in incorrect feature representation in the spectral domain. Since all three models were pre-trained using the ImageNet dataset, the training images were normalized using the mean and standard deviation of the ImageNet dataset~\cite{deng2009imagenet}.  The input images were resized accordingly to the specific dimensions for the respective architectures.\\

\subsection{Wavelet Feature Extraction for ConvNeXt}
The wavelet transform is shown to be effective for image feature extraction as it decomposes images into multi-resolution frequency components~\cite{mallat1989theory}. The ability to decompose an image into different frequency sub-bands while preserving spatial relationships makes it particularly effective at identifying edges, textures, and subtle patterns of an image~\cite{guo2022review}. Using the 2D Wavelet transform, an image can be decomposed into \emph{approximate} and \emph{detail} coefficients~\cite{porwik2004haar}. As such, we performed a single-level 2D Discrete Wavelet Transform (2D-DWT) using Haar wavelets to produce the approximate and detailed coefficients of the images and used the computed coefficients as features. 

\begin{figure}[t]
\centering
\includegraphics[scale=0.5]{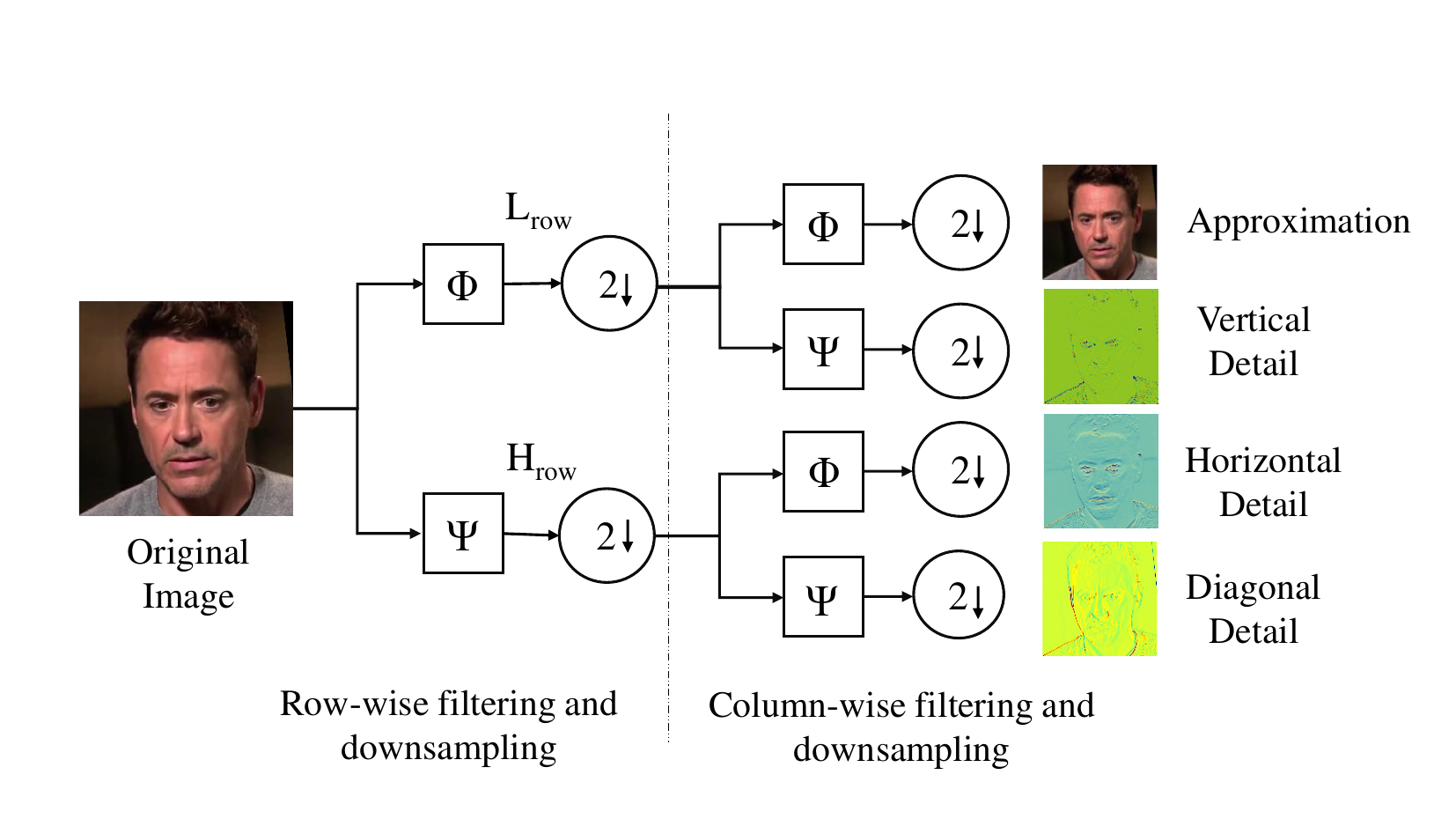}
\caption{Row- and column-wise filtering and downsampling to generate approximation and detail images through 2D-Discrete Wavelet Transform (DWT) using Haar wavelets.}
\label{figure_wavelet}
\end{figure}

\begin{figure}[t]
\centering
\includegraphics[scale=0.5]{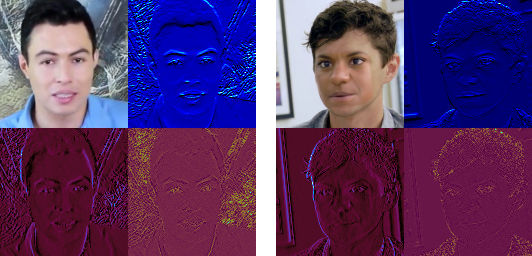}
\caption{Real and Fake feature images after wavelet transform. The top left, top right, bottom left, and bottom right portions indicate the approximate horizontal detail, vertical detail, and diagonal detail coefficients of the original image, respectively. Different colormaps are applied to the horizontal, vertical, and diagonal detail coefficients, and their contrasts are increased for better visibility.}
\label{figure_3}
\end{figure}

The 2D-DWT is applied to an image \( I(x, y) \) to decompose it into frequency sub-bands while preserving spatial information. Given an image \( I \in \mathbb{R}^{M \times N} \), a single-level 2D-DWT using Haar wavelets generates four coefficient matrices: the approximate coefficients and three sets of detail coefficients (horizontal, vertical, and diagonal). The wavelet decomposition is performed using separable filtering along rows and columns. The Haar wavelet uses two filters:
\begin{itemize}
    \item \textbf{Low-pass filter (\(\phi\))}: Coefficients \( \left[\frac{1}{\sqrt{2}}, \frac{1}{\sqrt{2}}\right] \)
    \item \textbf{High-pass filter (\(\psi\))}: Coefficients \( \left[-\frac{1}{\sqrt{2}}, \frac{1}{\sqrt{2}}\right] \)
\end{itemize}
As shown in Figure~\ref{figure_wavelet}, the sub-bands are computed as follows:
\begin{enumerate}
    \item Row-wise filtering and downsampling:
    \begin{align*}
        L_{\text{row}} &= \left( I \ast \phi \right) \downarrow_{2,\text{cols}} \\
        H_{\text{row}} &= \left( I \ast \psi \right) \downarrow_{2,\text{cols}}
    \end{align*}
    
    \item Column-wise filtering and downsampling:
    \begin{align*}
        A &= \left( L_{\text{row}} \ast \phi \right) \downarrow_{2,\text{rows}} \quad \text{(Approximation, LL)} \\
        V &= \left( L_{\text{row}} \ast \psi \right) \downarrow_{2,\text{rows}} \quad \text{(Vertical Details, LH)} \\
        H &= \left( H_{\text{row}} \ast \phi \right) \downarrow_{2,\text{rows}} \quad \text{(Horizontal Details, HL)} \\
        D &= \left( H_{\text{row}} \ast \psi \right) \downarrow_{2,\text{rows}} \quad \text{(Diagonal Details, HH)},
    \end{align*}
\end{enumerate}
where \( \ast \) indicates convolution; \( \downarrow_{2,\text{cols}} \), and \( \downarrow_{2,\text{rows}} \) indicate downsampling the columns and rows by 2, respectively; \( A \) denotes low-frequency approximation coefficients; and \( H, V, D \) denote high-frequency horizontal, vertical, and diagonal detail coefficients, respectively. For an RGB image \( I_r, I_g, I_b \) consisting of three color channels, we apply the above transformation independently to each color channel:
\begin{equation}
    A_c, H_c, V_c, D_c = \text{DWT}(I_c), \quad c \in \{r, g, b\},
\end{equation}
where \( A_c, H_c, V_c, D_c \) are the approximate and detail coefficients for each color channel \( c \). The final feature representation is formed by concatenating the coefficients across all three channels:
\begin{equation}
    F = \text{concat}((A_r, A_g, A_b), (H_r, H_g, H_b), (V_r, V_g, V_b), (D_r, D_g, D_b)).
\end{equation}
To prepare the wavelet-domain features for classification, the coefficients are scaled to an 8-bit integer range and resized to match the input dimensions of the ConvNeXt model. In Figure~\ref{figure_3}, we showed two such feature images with the approximate and detailed coefficients of two sample images from the dataset.

\subsection{Details of the Model Architectures}
Recall that our proposed solution utilizes three neural network models for a weighted ensemble approach for our final prediction. We chose each of these models to focus on different aspects of a deepfake image so that the ensemble approach provides better accuracy than individual models operating alone. The details of the models in our proposed solution are presented below.

\subsubsection{EfficientNet}

EfficientNet-B0 is a computationally efficient model whose architecture is optimized via neural architecture search (NAS) to maximize accuracy under a FLOP constraint, resulting in a highly efficient baseline that outperforms models with similar computational cost~\cite{tan2019efficientnet}. The core block of the EfficientNet architecture is the Mobile Inverted Bottleneck Convolution (MBConv) block inspired by MobileNetV2~\cite{sandler2018mobilenetv2}.
The MBConv block consists of a 1x1 Convolution, which expands the input channels to increase feature dimensionality. Then, a 3x3 or 5x5 depthwise convolution is applied to capture spatial features efficiently. Squeeze-and-Excitation (SE) recalibrates channel importance by computing a channel-wise attention vector~\cite{hoang2021practical}. The SE block acts as a lightweight attention mechanism to adaptively recalibrate each feature channel's importance~\cite{hu2018squeeze, xue2022partial}. It is responsible for two main operations:\\
\textbf{Squeeze:} Global spatial information is aggregated using global average pooling to generate channel-wise descriptors. Given an input feature map $X \in \mathbb{R}^{H \times W \times C}$, the squeezed feature vector $z \in \mathbb{R}^{C}$ is obtained by squeezing function $F_{sq}(.)$ as:
\begin{equation}
z_c = \frac{1}{H W} \sum_{i=1}^{H} \sum_{j=1}^{W} X_c(i,j), \quad c = 1, 2, ..., C,
\end{equation}
where $X_c(i,j)$ represents the activation at spatial location $(i,j)$ in channel $c$.\\

\noindent\textbf{Excitation:} The descriptors generated from the Squeeze block are passed through a lightweight two-layer fully connected network with a bottleneck structure to model non-linear channel independencies.\\
After that, a skip connection adds the input to the output if the input and output shapes match. The swish activation~\cite{ramachandran2017swish} function is applied after each convolution except the final projection.
\begin{equation}
\text{Swish}(x) = x \cdot \sigma(x) = \frac{x}{1+e^{-x}}
\end{equation}
EfficientNet-B0 uses MBConv1 (expansion factor 1) to MBConv6 (expansion factor 6) variants, with kernel sizes of 3x3 or 5x5 depending on the stage. A summary of different layers, the output shapes, and the number of parameters in the EfficientNet-B0 architecture is shown in Table~\ref{tab:efficientNetb0}.

\begin{table}[t]
    \centering
    \caption{Summary of the EfficientNetB0 model architecture}
    \resizebox{0.7\textwidth}{!}{
    \begin{tabular}{l l l}
    \hline 
    \textbf{Layer (type)} & \textbf{Output Shape} & \textbf{Param \#} \\
    \hline
    Conv2D + BatchNorm + SiLU & [32, 112, 112] & 928\\
    DepthwiseSeparableConv & [16, 112, 112] & 1,448 \\
    InvertedResidual-1 $\times$ 2 & [24, 56, 56] & 16,714 \\
    InvertedResidual-2 $\times$ 2  & [40, 28, 28] & 46,640\\
    InvertedResidual-3 $\times$ 3 & [80,14,14] & 242.930 \\
    InvertedResidual-4 $\times$ 3 & [112, 14, 14] & 543,148 \\
    InvertedResidual-5 $\times$ 4 & [192, 7, 7] & 2,026,348 \\
    InvertedResidual-6 & [320, 7, 7] & 717,232 \\
    Conv2d + BatchNorm + SiLU & [1280, 7, 7] & 412,160 \\
    AdaptiveAvgPool2d + Flatten & [1280] & 0\\
    Linear & [1] & 1,281\\
    \hline
    \multicolumn{2}{c} {Total Parameters} & 4,008,829 \\
    \multicolumn{2}{c} {Trainable Parameters} & 4,008,829 \\ 
    \multicolumn{2}{c} {Non-Trainable Parameters} & 0 \\
    \hline
    \end{tabular}}
    
    \label{tab:efficientNetb0}
\end{table}

\subsubsection{Data Efficient Image Transformer (DeiT)}
Unlike conventional Recurrent Neural Networks (RNNs), transformers process all inputs simultaneously, relying on a self-attention mechanism~\cite{xue2024self}. This allows transformers to learn long-range dependencies and global contexts of data~\cite{khan2022transformers}. The data-efficient image transformer, based on the backbone of the vision transformer, has been shown to work well in cases where training data is limited~\cite{touvron2021training}. Here, an image is divided into a grid of smaller patches as a sequence of input tokens ($N$ number of $16 \times 16$ patches), and each patch is associated with a vector through linear embedding. After adding positional encoding for each patch, the vectors are fed into the transformer encoder, where the self-attention mechanism helps to learn the relationship among all the patches~\cite{dosovitskiy2020image}. Note that the attention mechanism is based on a trainable associative memory of (key, value) pairs. A query vector $\mathbf{Q}$ is matched against key vectors $\{\mathbf{K}\}$ using dot products, scaled by $\sqrt{d}$, and normalized via a softmax function to produce weights. These weights are used to compute a weighted sum of value vectors $\{\mathbf{V}\}$, producing an output matrix as:
\begin{equation}
    \text{attention}\left(\mathbf{Q}, \mathbf{K}, \mathbf{V}\right) = \text{softmax}\left(\frac{\mathbf{Q} \mathbf{K}^\top}{\sqrt d}\right)\mathbf{V}
    \label{attention}
\end{equation}
In self-attention, the queries, keys, and values are derived from the same input sequence via linear transformations~\cite{vaswani2017attention}. Multi-head self-attention (MSA) applies $h$ separate attention functions (heads) in parallel. Each head provides a sequence of outputs, which are concatenated and projected back to the original dimension. This allows the model to attend to different parts of the input sequence simultaneously.

DeiT is an effective alternative to traditional CNNs in learning global features~\cite{khan2022transformers}, using knowledge distillation techniques and optimization strategies to improve generalization. DeiT learns from a pre-trained CNN model (that is, the RegNetY-16GF model~\cite{radosavovic2020designing}) and uses a distillation token during training, allowing it to learn faster and more efficiently with less data~\cite{touvron2021training}.

The DeiT architecture introduces a new hard distillation mechanism where they treat the hard decisions of the teacher model $y_t$ as true labels along with the actual true labels, $y$. If $Z_t$ are the logits of the teacher model and $Z_s$ are the logits of the student model, then the hard decision of the teacher model is defined as
\begin{equation}
y_t = \text{argmax}_c (Z_t(c))
\end{equation}
Here, $c$ is the class index. Using this prediction, the final loss function for hard distillation is treated as
\begin{equation}
L^{\text{hard dist}}_\text{global} = \frac{1}{2}L_{CE}(\text{softmax}(Z_s), y) + \frac{1}{2}L_{CE}(\text{softmax}(Z_s), y_t)
\end{equation}
The first term of the equation can be thought of as $L_\text{cross-entropy}$ and the second term as $L_\text{teacher}$.

Our proposed solution utilizes the DeiT-B model, a 768-dimensional model with 12 heads and 86 million parameters, based on the ViT-B model~\cite{touvron2021training}. The summary number of parameters and the output shapes of layers are shown in Table \ref{tab:DeiT}.

\begin{table}[t]
\centering
\caption{Summary of the data-efficient image transformer (DeiT) model architecture}
\resizebox{0.5\textwidth}{!}{
\begin{tabular}{l l l}

\hline
\textbf{Layer (type)} & \textbf{Output Shape} & \textbf{Param \#} \\ \hline
Conv2d & [764, 14, 14] & 590,592\\ 
Patch Embed & [196, 768] & 0\\ 
Dropout + Identity-1 & [197, 768] & 0\\
LayerNorm $\times$ 25 & [197, 768] & 38,400\\ 
Linear-1 $\times$ 12 & [197, 2304] & 21,261,312\\
Identity-2 & [12, 197, 64] & 0\\
(Linear-2 + Attention) $\times$ 12 & [197, 768] & 35,407,872\\ 
(Linear-3 + GELU) $\times$ 12 & [197, 3072] & 28,348,416\\ 
Dropout & [197, 3072] & 0 \\
Mlp $\times$ 12 & [197, 768] & 0\\
Identity and Dropout & [768] & 0\\
Linear-4 & [2] & 1,538 \\
Vision Transformer & [2] & 0\\ \hline
\multicolumn{2}{c} {Total Parameters} & 85,648,130 \\
\multicolumn{2}{c} {Trainable Parameters} & 85,648,130 \\ 
\multicolumn{2}{c} {Non-Trainable Parameters} & 0
\\ \hline
\end{tabular}}
\label{tab:DeiT}
\end{table}

%%%%%%%%%%%% proposed ensemble figure %%%%%%%%%%%%%
\begin{figure*}[ht]
\centering
\includegraphics[scale=0.38]{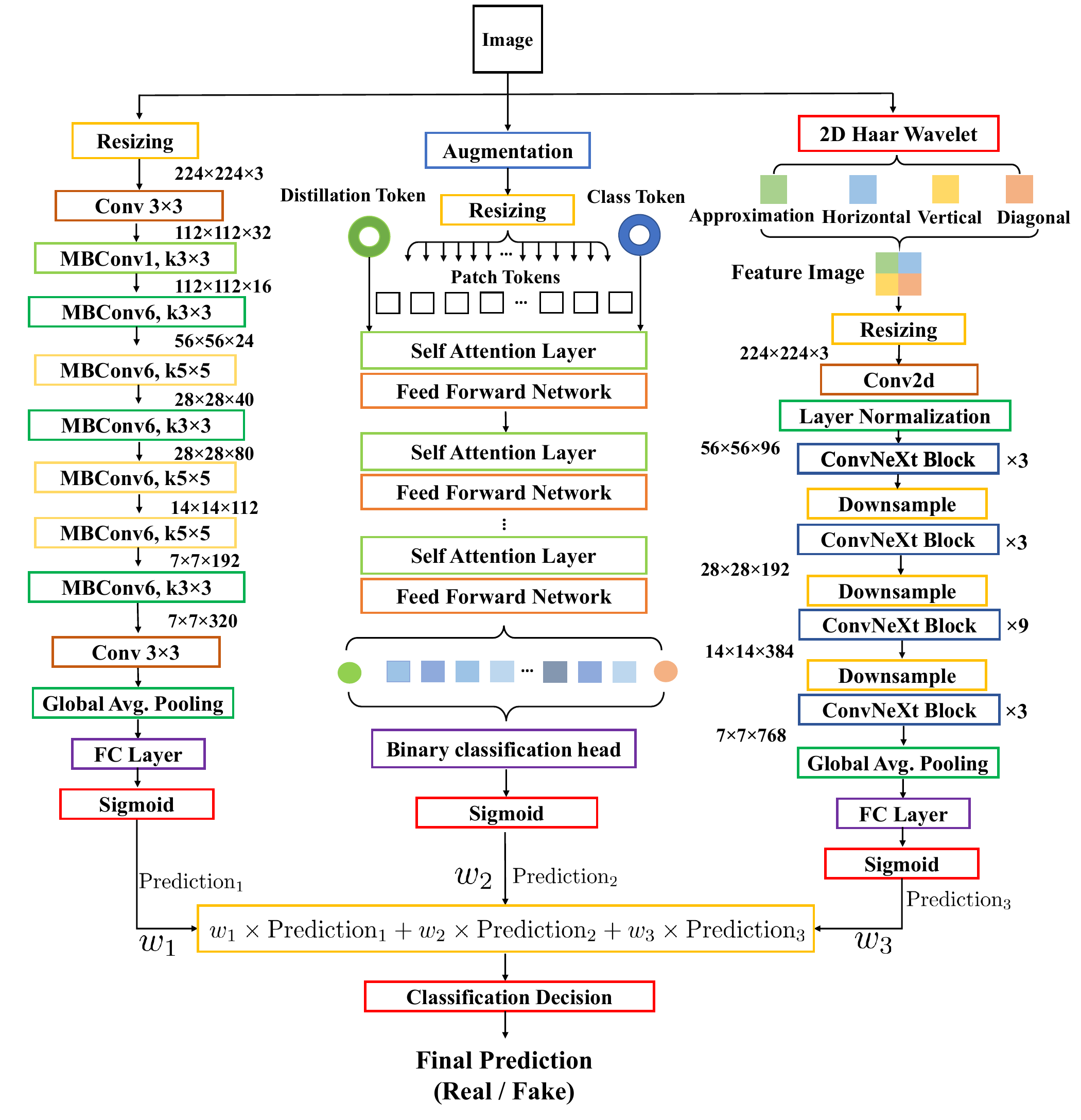}
\caption{Detail of the proposed CAE-Net architecture with weighted ensemble framework.}
\label{figure_9}
\end{figure*}
%%%%%%%%%%%% proposed ensemble figure %%%%%%%%%%%%%

\subsubsection{ConvNeXt Model}

ConvNeXt combines the efficiency and inductive biases of traditional CNNs with the scalability and performance of transformers. It employs an inverted bottleneck block, similar to transformers' MLP blocks, reducing FLOPs in downsampling layers. Here, batch normalization is replaced with layer normalization, aligning with transformer practices, and GELU is used instead of ReLU for smoother non-linearity~\cite{liu2022convnet}.\\
Inspired by both ResNets~\cite{he2016deep} and Swin transformers~\cite{Liu_2021_ICCV}, the architecture includes a patchify stem where input images are processed by a 4x4 convolution with stride 4, followed by layer normalization, to downsample the image into non-overlapping patches. This mimics ViT’s patch embedding but uses convolutions.\\
Each ConvNeXt block contains a 7x7 depthwise convolution (one filter per input channel) that is applied to capture spatial features efficiently. Then, layer normalization is applied before and after the convolution to stabilize training. It normalizes features across the channel dimension for each spatial position. For feature $ X[i,j,:]~ \epsilon~ \mathbf{R^C} $ the normalized output is:
    \begin{equation}
        \hat{X}[i,j,c] = \frac{X[i,j,c]-\mu}{\sqrt{\sigma^2+\epsilon}}\cdot \gamma + \beta
    \end{equation}
Here $\mu = \frac{1}{C}\sum_{c = 1}^C X[i,j,c]$, $\sigma^2 = \frac{1}{C}\sum_{c = 1}^C (X[i,j,c] - \mu)^2 $ , $\epsilon$ is a small constant, and $\gamma$ and $\beta$ are trainable parameters. This type of normalization works better than typical batch normalization~\cite{ba2016layer}.
After that, pointwise convolution is applied to expand (to 4x the input dimension) and contract the feature map, mimicking transformer MLP layers. GELU (Gaussian Error Linear Unit)~\cite{hendrycks2016gaussian} is applied instead of RELU for smoother non-linearity.
    \begin{equation}
    \text{GELU}(x) = x\cdot \Phi(x) = x \cdot \frac{1}{2}[1+\text{erf}(\frac{x}{\sqrt{2}})]
    \end{equation}
A skip connection is used from the input to the block's output, similar to ResNet's residual connections~\cite{he2016deep}. Between stages, ConvNeXt uses a downsampling layer with layer normalization and a 2x2 convolution with a stride of 2, inspired by Swin Transformer’s separate downsampling strategy, unlike ResNet’s within-block downsampling. From the diverse family of ConvNeXt architectures, we specifically implemented the ConvNeXt-tiny model for its low compute and high accuracy on the dataset. A summary of the different layers in the ConvNeXt-tiny architecture that we have implemented is presented in Table~\ref{ConvNeXt_tiny}.

\begin{table}[htbp]
    \centering
    \caption{Summary of the ConvNeXt architecture}
    \resizebox{0.7\textwidth}{!}{
    \begin{tabular}{lll}
    \hline
        \textbf{Layer Type} & \textbf{Output Shape} & \textbf{Param \#} \\
    \hline
    Conv2d + LayerNorm2d & $[96, 74, 74]$ & 4,896 \\
    CNBlock ($\times 3$) & $[96, 74, 74]$ & 237,600 \\
    LayerNorm2d + Conv2d  & $[192, 37, 37]$ & 74,112 \\
    CNBlock ($\times 3$) & $[192, 37, 37]$ & 917,568 \\
    LayerNorm2d + Conv2d & $[384, 18, 18]$ & 295,680 \\
    CNBlock ($\times 9$) & $[384, 18, 18]$ & 10,813,824 \\
    LayerNorm2d + Conv2d & $[768, 9, 9]$ & 1,181,184 \\
    CNBlock ($\times 3$) & $[768, 9, 9]$ & 14,287,104 \\
    AdaptiveAvgPool2d + LayerNorm2d & $[768, 1, 1]$ & 1536\\
    Flatten + Linear & $[1]$ & 769 \\
    \hline
    \multicolumn{2}{c} {Total Parameters} & 27,814,273 \\
    \multicolumn{2}{c} {Trainable Parameters} & 27,814,273 \\ 
    \multicolumn{2}{c} {Non-Trainable Parameters} & 0\\
\hline
    \end{tabular}}
    \label{ConvNeXt_tiny}
\end{table}

\subsubsection{Proposed Weighted Ensemble Architecture: CAE-Net}
Finally, we implemented a weighted ensemble mechanism to combine the predictions of the aforementioned three models. The associated weights for the EfficientNet, DeiT, and ConvNeXt with wavelet transform are 0.35, 0.34, and 0.31, respectively, based on our parameter search. Initially, each model was given equal weights (accuracy result for this selection of weights is stated in Table~\ref{tab: Ensemble ablation}) and then manually tuned. To refine the fusion strategy, the weights were then manually tuned based on validation performance. Specifically, weight combinations were evaluated in 0.01 increments while satisfying the simplex constraint ($w_1 + w_2 + w_3 = 1$), and the configuration that achieved the highest accuracy on the SP Cup 2025 validation set was selected. This ensemble employs probability-level late fusion, where each model processes the input independently within its full architecture, preserving its native feature representation. The final architecture of CAE-Net is shown in Figure \ref{figure_9}.

\subsection{Adversarial Perturbations}
Adversarial perturbations are crafted noises or modifications added to classifier model inputs, often imperceptible to humans, that can significantly deteriorate model performance and lead to misclassification. Such adversarial perturbations can be universal or image-dependent~\cite{poursaeed2018generative} and are usually done in white-box settings where access to the model's architecture and parameters is required. Also, adversarial attacks on a known network have been shown to work effectively against other networks as well~\cite{goodfellow2014explaining}, making black-box attacks an imminent threat for sensitive issues like deepfake detection~\cite{Neekhara_2021_CVPR}. We evaluate the robustness of our proposed models by testing them against a simple but effective adversarial perturbation, the Fast Sign Gradient Method (FGSM). FGSM is a white-box attack without specifying a target class. Using this method, adversarial images are created by calculating the gradient of loss for input data and adjusting the input by a small amount in the direction of the gradient sign~\cite{goodfellow2014explaining}.
\begin{equation}
    x_\text{adv} = x + \epsilon \cdot \text{sign}(\nabla_xJ(\theta,x,y))
\end{equation}
Here $(x,y)$ are the original input image and its label, $\theta$ indicates model parameters, $x_{adv}$ is the adversarially perturbed image, and $\nabla_xJ(\theta,x,y)$ is the gradient of the loss function $J$ with respect to the input image $x$. The parameter $\epsilon$ controls the magnitude of the perturbation added to the original image. This method works like a gradient ascent algorithm trying to maximize the loss with the smallest possible perturbation, $\epsilon$, which makes it barely noticeable to the human eye. Though FGSM is one of the simplest forms of perturbations, it has been shown to reduce the accuracy of high-quality detectors to a large extent~\cite{9207034}. Its effectiveness for binary classification problems makes it an ideal choice for testing the robustness of our proposed method.

\section{Experimental Results}
\label{sec:result}
In this section, we demonstrate the performance of our proposed solution on the competition dataset. No prior work exists on this novel combined deepfake image detection dataset, which consists of images from eight different benchmark deepfake video datasets. To address this, we hypothesized that lightweight CNNs(e.g., EfficientNet) would excel in resource-constrained settings, modern CNNs (e.g., ConvNeXt) would leverage transformer-inspired designs for better feature extraction, and ViTs would capture the global context of images. We systematically evaluated these models to test these hypotheses along with multiple other frequency domain feature extraction techniques, establishing the first performance benchmarks for the dataset. We note that all the results noted for different benchmark models in Table~\ref{tab:CNN experimentations},~\ref{tab:transformer experimentations},~\ref{tab:frequency domain experimentations} are done with a fixed random seed of 42.\\

\begin{table}[htbp]
    \centering
    \caption{Comparative analysis of different CNN model variations to capture local artifacts}
    \label{tab:CNN experimentations}
    \resizebox{0.55\textwidth}{!}{
    \begin{tabular}{l c r}
        \hline
        \textbf{Classifier Models} & \textbf{Accuracy} & \textbf{\# of params.} \\
        \hline
        VGG-19~\cite{simonyan2014very} & 62.25\% & 144 M\\
        ResNet-50~\cite{he2016deep} & 69.00\% & 25.6 M\\
        MobileNet-V2~\cite{sandler2018mobilenetv2} & 78.12\% & 3.5 M\\
        ResNet-34~\cite{he2016deep} & 80.32\% & 21.3 M\\
        RegNet-400mF~\cite{9156494} & 80.92\% & 5.2 M\\
        Inception V3~\cite{Szegedy_2016_CVPR} & 84.38\% & 26 M\\
        ResNet-34 with CBAM\tablefootnote{Convolutional Block Attention Module} & 85.19\% & 24.5 M\\
        ResNet-34 with SE\tablefootnote{Squeeze and Excitation}  & 88.90\% & 21.3 M\\
        DenseNet-121~\cite{huang2017densely} & 90.01\% & 7.98 M \\
        \textbf{EfficientNet-B0}~\cite{tan2019efficientnet} & \textbf{90.79\%} & 3.96 M\\
        EfficientNet-B4~\cite{tan2019efficientnet} & 87.27\% & 19.3 M\\
        EfficientNet-B5~\cite{tan2019efficientnet} & 88.31\% & 30.4 M\\
        ResNeXt-50~\cite{Xie_2017_CVPR} & 85.19\% & 25 M\\
        ConvNeXt~\cite{liu2022convnet} & 88.18\% & 27.8 M\\
        \hline
    \end{tabular}
    }
\end{table}
The performance and size comparison of different benchmark CNN models are shown in Table~\ref{tab:CNN experimentations}. VGG-19 is the largest model by a significant margin, but has the lowest accuracy. The use of attention mechanisms like CBAM and SE blocks in the ResNet-34 variants appears beneficial, as the 88.90\% accuracy of ResNet-34 with SE is substantially higher than the baseline ResNet-34 (80.32\%). This led to our choice of EfficientNet, which has in-built SE blocks and is less compute-heavy compared to other CNNs. Among different variants of EfficientNets,  EfficientNet-B0 has the highest performance with the lowest number of parameters, making it an optimum choice for a backbone model capturing local artifacts. DenseNet and ConvNeXt show high performance as well, making them good candidates for the frequency domain experimentations.
\begin{table}[ht]
    \centering
    \caption{Comparative analysis of different transformer model variations to capture global artifacts}
    \label{tab:transformer experimentations}
    \resizebox{0.55\textwidth}{!}{
    \begin{tabular}{l c r}
        \hline
        \textbf{Classifier Models} & \textbf{Accuracy} & \textbf{\# of params.} \\
        \hline
        ViT-B~\cite{sharir2021imageworth16x16words} & 78.12\% & 86.6 M\\
        Swin Transformer~\cite{Liu_2021_ICCV} & 86.52\% & 28.3 M\\ MobileViT~\cite{mehta2022mobilevitlightweightgeneralpurposemobilefriendly} & 87.04\% & 5.6 M\\
        \textbf{DeiT-B}~\cite{touvron2021training} & \textbf{89.32\%} & 85.6 M\\
        \hline
    \end{tabular}
    }
\end{table}

As shown in Table~\ref{tab:transformer experimentations}, the ViT-B and DeiT-B models are the largest, both with parameter counts around 86 M. Despite being similar in size, DeiT-B significantly outperforms ViT-B in accuracy, suggesting the training strategies used in DeiT are highly effective. Among the transformers, MobileViT is the most parameter-efficient model using only 5.6 M parameters. However, since achieving the highest accuracy is of the utmost importance in sensitive issues like deepfake detection, we take DeiT-B as a backbone model to capture global artifacts.

\begin{table}[htbp]
    \centering
    \caption{Comparative analysis of different CNN model variations to capture frequency domain artifacts}
    \label{tab:frequency domain experimentations}
    \resizebox{0.60\textwidth}{!}{
    \begin{tabular}{l c r}
        \hline
        \textbf{Classifier Models} & \textbf{Accuracy} & \textbf{\# of params.} \\
        \hline
        ResNet-34 with FFT~\cite{cooley1965algorithm} & 66.55\% & 21.3 M\\
        ResNet-34 with DCT~\cite{ahmed2006discrete}  & 71.48\% & 21.3 M\\
        ResNet-34 with Wavelet~\cite{mallat1989theory} & 80.44\% & 21.3 M\\
        EfficientNetB0~\cite{tan2019efficientnet} with Wavelet & 82.48\% & 3.96 M\\
        RegNet-Y~\cite{9156494} with Wavelet & 86.04\% & 10.31 M\\
        Xception-Net~\cite{chollet2017xception} with Wavelet & 87.76\% & 20.8 M\\
        \textbf{ConvNeXt}~\cite{liu2022convnet} \textbf{with Wavelet} & \textbf{89.49\%} & 27.81 M\\
        \hline
    \end{tabular}
    }
\end{table}

Additionally, we conducted a comparative analysis of different frequency domain feature extraction techniques, such as DCT, Fast Fourier Transform (FFT), and 2D Discrete Wavelet Transform (DWT), as preprocessing steps with benchmark CNN models and the results are shown in Table \ref{tab:frequency domain experimentations}. DCT and FFT with ResNet-34 yield very low accuracy among the specialized models. In contrast, ResNet-34 with Wavelet significantly improves performance to 80.44\%, suggesting that the Wavelet transform is generally more effective than the DCT for this specific classification task. Therefore, we experimented with other high-performing CNN models with the wavelet transform as a pre-processing technique. The ConvNeXt with Wavelet model achieves the highest accuracy at 89.49\%. Therefore, we take this model as a backbone for capturing frequency domain artifacts specifically.\\

\subsection{Choice of Hyperparameters}
Due to computational resource constraints, a comprehensive hyperparameter tuning using advanced optimization algorithms was not feasible. Instead, we conducted a discrete evaluation of key hyperparameters to identify optimal settings for the first stage of our five-stage training process, which spanned five epochs. We assessed batch sizes of 16, 32, and 64 (each with a fixed learning rate of $1\times10^{-4}$) and learning rates of $1\times10^{-3}$, $1\times10^{-4}$, and $1\times10^{-5}$ (each with a fixed batch size of 32). This analysis focused on recording the validation accuracy of EfficientNet, DeiT, and ConvNeXt for each configuration. The results are presented in Table~\ref{tab: Hyperparameter configurations}, which guided the selection of the most effective settings, balancing performance and resource efficiency. We acknowledge this as a limitation and plan to explore more extensive tuning in future work with adequate resources.  

\begin{table}[htbp]
    \centering
    \caption{Accuracy results for different hyperparameter configurations}
    \label{tab: Hyperparameter configurations}
    \begin{tabular}{l c c c}
    \hline
    \multirow{2}{*}{Parameter} & \multicolumn{3}{c}{Accuracy} \\
   \cline{2-4}
    & EfficientNet & DeiT & ConvNeXt\\
    \hline
    Batch Size: & & & \\
    ~~~~~~~16 & \textbf{91.40\%} & \textbf{89.78\%} & 87.37\%\\
    ~~~~~~~32 & 90.79\% & 89.32\% & 89.49\%\\
    ~~~~~~~64 & 87.34\% & 84.86\% & \textbf{91.41\%}\\
    Learning Rate: & & & \\
    ~~~~~~1$\times 10^{-3}$ & 84.96\% & 49.61\% & 50.39\%\\
    ~~~~~~1$\times 10^{-4}$ & \textbf{90.79\%} & 89.32\% & \textbf{89.49\%}\\
    ~~~~~~1$\times 10^{-5}$ & 76.95\% & \textbf{90.79\%} & 87.66\%\\
    \hline
    \end{tabular}
\end{table}

As shown in Table~\ref{tab: Hyperparameter configurations}, EfficientNet and DeiT achieved the highest accuracy for a batch size of 16, while ConvNeXt performed the best with a batch size of 64. For learning rate selection, the CNN models performed well at $1\times10^{-4}$, whereas DeiT achieved its best result at $1\times10^{-5}$. Considering these observations, we selected a learning rate of $1\times10^{-4}$ with a batch size of 32 for ensemble training, providing a balanced configuration that supports stable convergence across all three models and results in improved ensemble accuracy.

For training, we used the ADAM optimizer with binary categorical cross-entropy as the loss function. All experiments were conducted on an NVIDIA Tesla P100 GPU. The final hyperparameters used for model training are summarized in Table~\ref{tab:Hyperparameters}. We have implemented similar hyperparameter settings for each stage in our five-stage training approach. Each stage was trained with an equal number of epochs to ensure consistency, and the images of both classes went under augmentation only while training our DeiT architecture.  

\begin{table}[htbp]
    \centering
    \caption{Hyperparameters for model training}
    \label{tab:Hyperparameters}
    \resizebox{0.5\textwidth}{!}{
    \begin{tabular}{lc}
        \hline
        \textbf{Hyperparameters} & \textbf{Details} \\
        \hline
        Batch size & 32 \\
        Number of Epochs in Each Stage & -- \\
        \quad EfficientNet & 5 \\
        \quad DeiT & 5 \\
        \quad ConvNeXt & 4 \\
        Learning rate & -- \\
        \quad Learning rate scheduler & ReducedLROnPlateau \\
        \quad Initial learning rate & 0.0001 \\
        \quad Reduction factor & 0.1 \\
        \quad Patience & 3 \\
        \hline
    \end{tabular}}
\end{table}

\subsection{Ablation study}

\begin{table*}[htbp]
    \centering
    \caption{Ablation study on ensemble for best performing models}
    \label{tab: Ensemble ablation}
    \resizebox{1\textwidth}{!}{
    \begin{tabular}{c c c c c c c c c}
    \hline
     \multirow{2}{*}{EfficientNet}  &  \multirow{2}{*}{DeiT} &  \multirow{2}{*}{ConvNeXt} &  \multirow{2}{*}{Ensemble} & \# of params & FLOPs & Inference & Throughput & Accuracy\\
     & & & & (M) & (G) & time (ms) & (image/s) & (\%)\\
     
     \hline
    \ding{51} & \ding{55} & \ding{55} & \ding{55} & 3.96  & 0.386 & 24.42 & 1310.26 & 86.11 $\pm$ 3.22 \\
    \ding{55} & \ding{51} & \ding{55} & \ding{55} &  85.64 & 16.863 & 178.57 & 179.2 & 89.16 $\pm$ 1.26\\
    \ding{55} & \ding{55} & \ding{51} & \ding{55} & 27.81 & 7.521 & 128.81 & 248.43 &  91.24 $\pm$ 1.22\\
    \ding{51} & \ding{51} & \ding{55} & equal weights & 89.61 & 17.247 & 202.95 &  157.67 & 92.59 $\pm$ 1.87\\
    \ding{51} & \ding{55} & \ding{51} & equal weights & 31.77 & 7.907 & 153.11 & 209.0 & 93.14 $\pm$ 1.43\\
    \ding{55} & \ding{51} & \ding{51} & equal weights & 113.46 & 24.384 & 307.64 & 104.02 & 91.87 $\pm$ 0.78\\
    \ding{51} & \ding{51} & \ding{51} & equal weights & 117.426 & 24.768 & 332.08 & 96.42 & 94.16 $\pm$ 0.61 \\
    \ding{51} & \ding{51} & \ding{51} & optimized weights & 117.426 & 24.768 & 332.08 & 96.42 & \textbf{94.46 $\pm$ 0.58} \\
    \ding{51} & \ding{51} & \ding{51} & majority voting & 117.426 & 24.768 & 332.08 & 96.42 & 94.09 $\pm$ 0.53 \\
     \hline    
    \end{tabular}}

\end{table*}
\noindent\textbf{Ablation study on ensemble configurations.} To better understand the individual contributions of each sub-module in the proposed ensemble, we conducted a detailed ablation study using the three backbone models, namely the EfficientNet, DeiT, and ConvNeXt, across various combinations. The analysis aimed to examine not only the detection performance but also the practical deployability of each configuration by reporting the number of parameters, floating point operations (FLOPs), inference time, and throughput (measured with a batch size of 32). Three ensemble strategies were evaluated: (i) equal weights, where all models contributed equally; (ii) optimized weights, where coefficients were manually tuned on the fixed validation set; and (iii) majority voting, which relied on discrete decision fusion. Each setting was trained and tested five times with random initializations, and the average accuracy with standard deviation is summarized in Table~\ref{tab: Ensemble ablation}. From the results, the individual models showed competitive but varying performance, with ConvNeXt achieving the highest single-model accuracy (91.24\%) among the three. Combining multiple models consistently improved results, demonstrating the complementary strengths of spatial (EfficientNet), global (DeiT), and frequency-domain (ConvNeXt + wavelet) features. The optimized weighted ensemble achieved the best overall accuracy (94.46\%), outperforming both equal weighting and majority voting approaches. In addition to performance, we included deployment-oriented metrics, such as parameter count, FLOPs, throughput, and inference time, to provide a clearer view of the trade-off between accuracy and computational cost. These added measurements help assess the real-world feasibility of the model, showing that the proposed ensemble maintains strong efficiency while offering significant accuracy gains over its individual components.\\

\begin{figure*}[ht] 
    \centering
    \begin{subfigure}{0.5\textwidth}
        \centering
        \includegraphics[width=\textwidth]{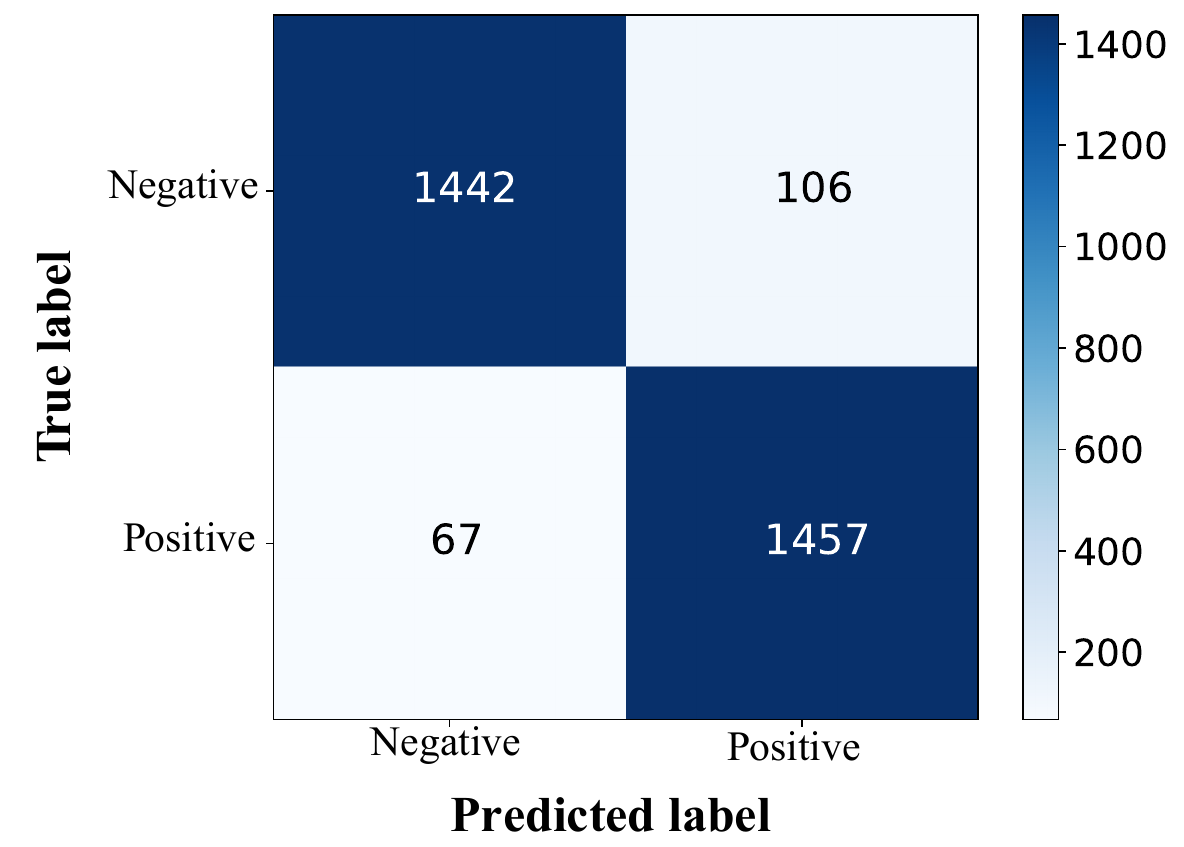}
        \caption{}
        \label{confusion matrix}
    \end{subfigure}
    % \hfill
    \begin{subfigure}{0.45\textwidth}
        \centering
        \includegraphics[width=\textwidth]{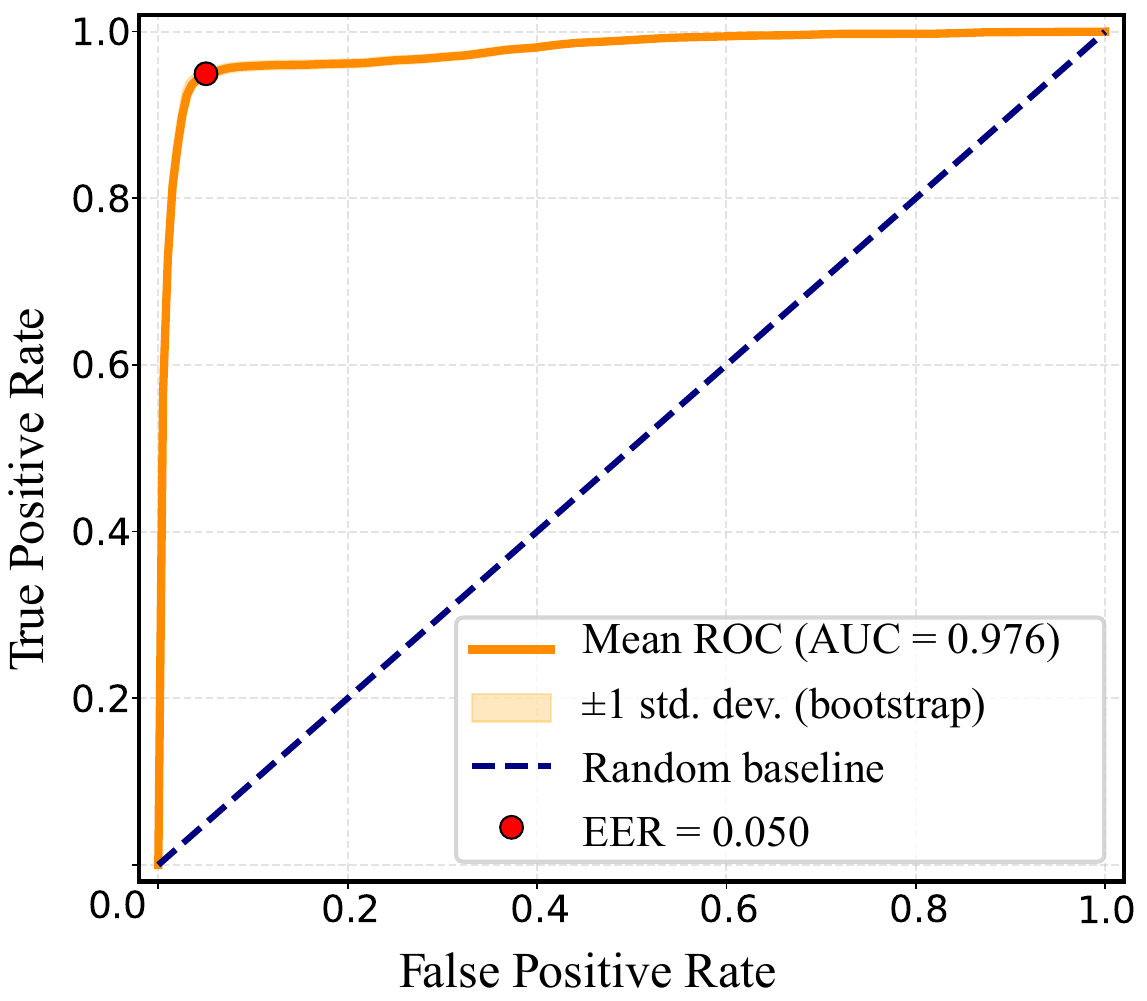}
        \caption{}
        \label{ROC curve}
    \end{subfigure}
    \caption{(a) Confusion matrix and (b) ROC curve of the proposed CAE-Net evaluated on the SP Cup 2025 validation set.}
    \label{fig:Ensemble_valid_metrics}
\end{figure*}

\noindent\textbf{Result Analysis.} We evaluated our weighted ensemble model for multiple evaluation metrics, including accuracy, precision, recall, F1-score, area under the receiver operating characteristic (AU-ROC) curve, equal error rate, and average precision. While calculating these metrics, real images were considered as class 0 and fake images as class 1. The performance of our model on these metrics is presented in Table \ref{tab:final results}, and the confusion matrix and the ROC curve are shown in Figure \ref{fig:Ensemble_valid_metrics}. The results shown here are for a fixed seed of 42. It is evident from the evaluation indices that our proposed weighted ensemble architecture performs very well on the competition dataset, which underlines the generalization capability of the proposed approach. \\
\begin{table}[htbp]
    \centering
    \caption{Performance of the final ensemble architecture combining ConvNext, DeiT, and EfficientNet}
    \label{tab:final results}
    \begin{tabular}{lc}
        \hline
        \textbf{Evaluation Metrics} & \textbf{Performance}\\
        \hline
        Accuracy & 94.46\% \\
        Precision & 93.70\%\\
        Recall &  95.60\% \\
        F1 score &  94.64\%\\
        Area under ROC & 97.60\%\\
        Equal Error Rate & 5.00\% \\
        Average Precision &  97.71\%\\
        \hline
    \end{tabular}
\end{table}
\subsection{Interpretability analysis of the proposed network}
\begin{figure*}[ht]
\centering
\includegraphics[scale=0.5]{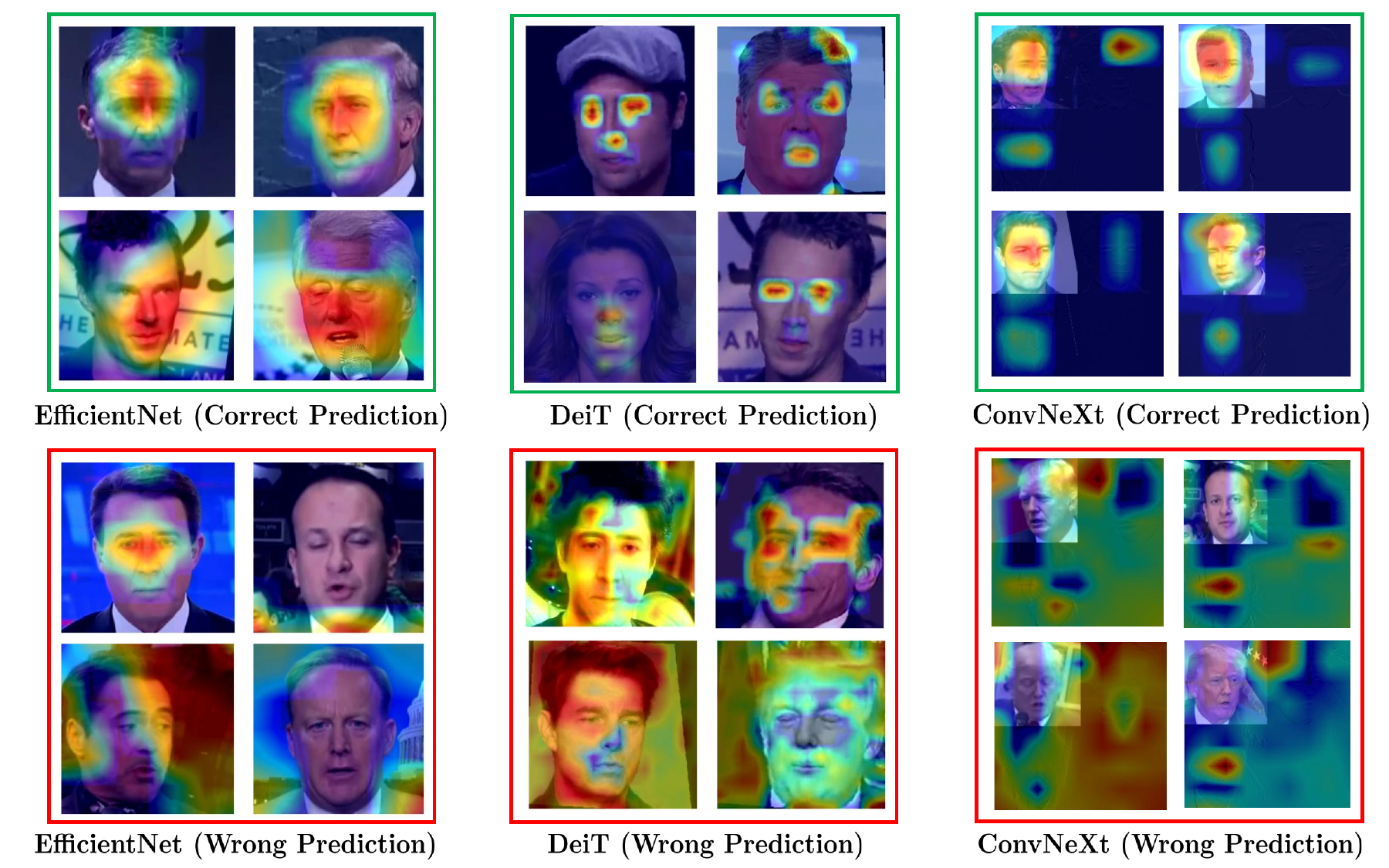}
\caption{Importance map shown by Grad-CAM for different models with correct and wrong predictions. The green boxes indicate the grad-CAM view for correct predictions, and the red boxes indicate the wrong ones.}
\label{fig:allfigures_grad}
\end{figure*}

To investigate the performance of our ensemble model, we intend to visualize i) the regions of the images that the three models focus on for classification, and ii) how the three models distinguish real and fake images before and after training. To that end, we note that the Grad-CAM visualizations show how different models focus on different facial regions for deepfake detection. In Figure~\ref{fig:allfigures_grad}, it can be seen that EfficientNet focuses on localized facial features, whereas DeiT displays diffuse attention patterns indicating a focus on global features. When the diverse insights from each model are combined into an ensemble, the overall prediction becomes more accurate, robust, and better at generalizing to new data. Additionally, we present the Grad-CAM view of some of the wrong predictions by the three models in Figure~\ref{fig:allfigures_grad}, showing the focus of irrelevant regions, leading to misclassifications.
%%%%%%%%%%%% Grad_CAM %%%%%%%%%%%%%

\begin{figure*}[ht]
\centering
\includegraphics[scale=0.45]{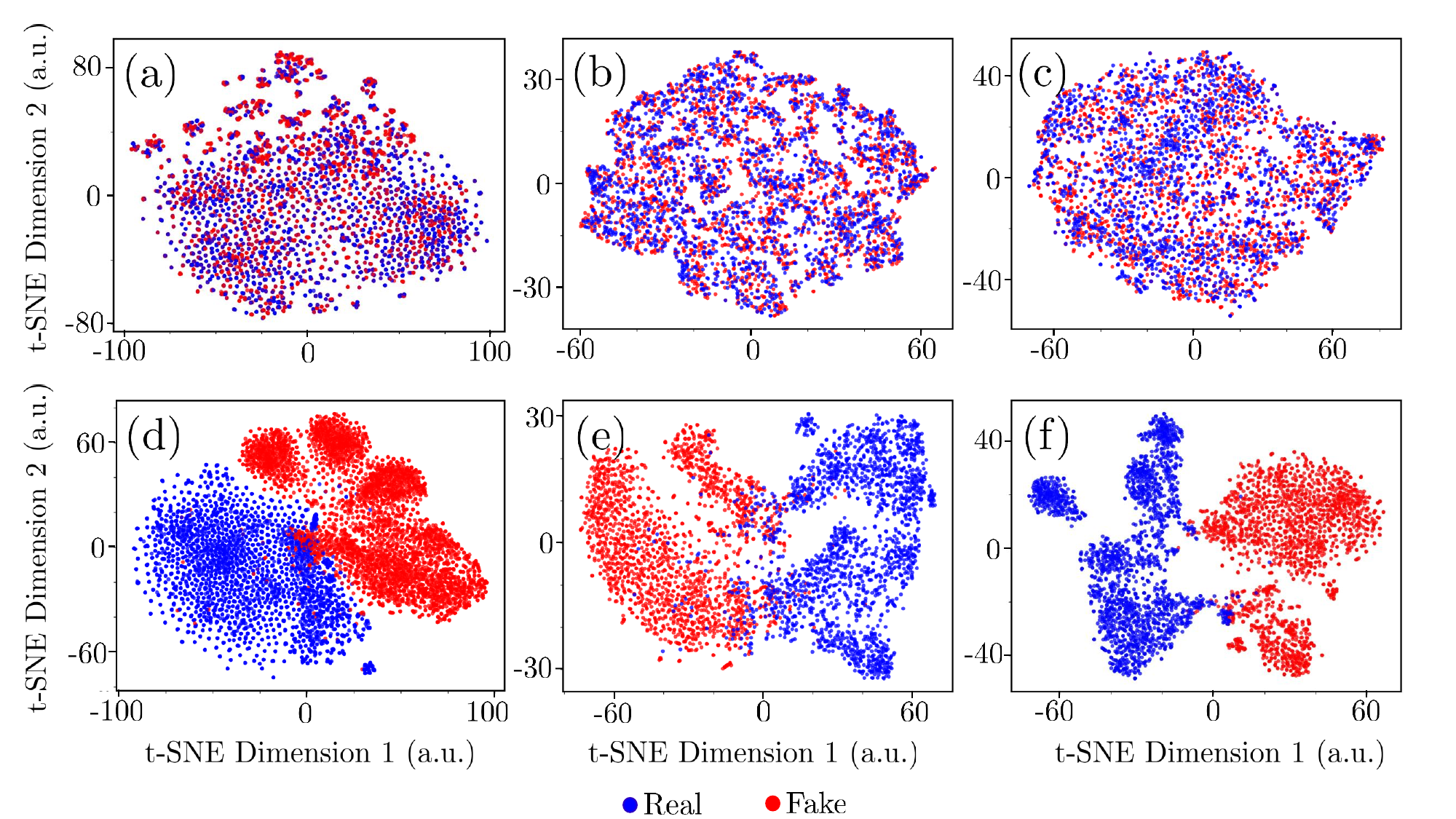}
\caption{t-SNE plot of three models before and after training for randomly selected 4000 real and fake images using perplexity = 40. (a) EfficientNet (Untrained), (b) DeiT (Untrained), (c) ConvNeXt (Untrained), (d) EfficientNet (Trained), (e) DeiT (Trained), and (f) ConvNeXt (Trained).}
\label{fig:allfigures_tSNE}
\end{figure*}

Next, we note that high-dimensional data can be visualized by projecting onto a two-dimensional space using the non-linear dimensionality reduction technique known as t-SNE (t-Distributed Stochastic Neighbor Embedding)~\cite{van2008visualizing}. Maintaining the local neighborhoods of data points shows the differences and similarities of data points in the learned feature space. As such, the t-SNE visualizations (Figure \ref{fig:allfigures_tSNE}) offer valuable insights into the effectiveness of our models. For both real and fake images, we observe cohesive clusters for all three models, despite the images being produced using various deepfake generation methods. This highlights the generalization capability of our proposed ensemble model. After training, all three backbone models, i.e., EfficientNet, DeiT, and ConvNeXt, produced well-separated embeddings for real and fake images, as illustrated in Figure \ref{fig:allfigures_tSNE}(d), (e), (f).  

%%%%%%%%%%%% t-SNE %%%%%%%%%%%%%
To quantitatively assess the cluster separability of real and fake embeddings in the t-SNE space, four widely used metrics were considered: Calinski–Harabasz (CH) Index~\cite{calinski1974dendrite}, Davies–Bouldin (DB) Index~\cite{4766909}, Inter-Centroid Euclidean Distance, and Inter-Centroid Mahalanobis Distance~\cite{mahalanobis2018generalized}. The CH Index measures the ratio of between-cluster dispersion to within-cluster compactness—higher values indicate better separation. The DB Index reflects average cluster similarity, where lower values denote reduced overlap. The Inter-Centroid Euclidean Distance quantifies how far apart the mean feature representations of two classes are, while the Mahalanobis Distance accounts for covariance, offering a more distribution-aware measure of separation.
\begin{table*}[htbp]
\centering
\caption{Cluster separability metrics (t-SNE space) for different models before and after training.}
\label{tab:tsne_separability_metrics}
\resizebox{1.0\textwidth}{!}{
\begin{tabular}{lcccc}
\hline
\textbf{Model} & Calinski--Harabasz $\uparrow$ & Davies--Bouldin $\downarrow$ & Inter-Centroid (Euclidean) $\uparrow$ & Inter-Centroid (Mahalanobis) $\uparrow$ \\
\hline
EfficientNet (Untrained) & 10.9841   & 16.3122 & 1.0779  & 1.9434 \\
EfficientNet (Trained)   & 358.1060  & 2.8366  & 2.5301  & 1.9628 \\
DeiT (Untrained)         & 2.5970    & 36.9521 & 1.9842  & 0.0643 \\
DeiT (Trained)           & 6799.6961 & 0.7083  & 68.3502 & 1.7020 \\
ConvNeXt (Untrained)     & 4.5949    & 27.3898 & 2.9342  & 0.1013 \\
ConvNeXt (Trained)       & 4530.9877 & 0.8663  & 61.1882 & 1.7710 \\
\hline
\end{tabular}}
\end{table*}

As shown in Table~\ref{tab:tsne_separability_metrics}, all three models demonstrate substantial improvements in separability after fine-tuning. For example, the Calinski–Harabasz score rises from 10.98 to 358.10 for EfficientNet, from 2.60 to 6799.70 for DeiT, and from 4.59 to 4530.99 for ConvNeXt, indicating a remarkable increase in inter-class distinction. Correspondingly, the Davies–Bouldin index decreases sharply across all models, confirming tighter and more compact clusters. Both Euclidean and Mahalanobis distances between class centroids also increase, showing that fine-tuning pushes the real and fake distributions farther apart in feature space. Together, these results provide strong quantitative evidence that the proposed training strategy significantly enhances feature discriminability, leading to more distinct and separable clusters in the t-SNE representation.

\subsection{Statistical analysis}

To further validate the effectiveness of the proposed ensemble model, we conducted a pairwise statistical comparison using the McNemar test. This non-parametric test assesses whether two classifiers differ significantly in their performance on paired nominal data, focusing specifically on instances where the models disagree. The results of this analysis are summarized in Table~\ref{tab:Mc-Nemar Statistical Analysis}.

Among the individual models, EfficientNet shows slightly better performance than ConvNeXt ($p=0.047$), with 40 more correct predictions in disagreement cases. However, this difference does not meet the Bonferroni-corrected significance threshold ($\alpha=0.008$). A statistically significant difference is observed between EfficientNet and DeiT ($p=0.003$), where EfficientNet outperforms DeiT by 58 correct predictions in disagreement cases. In contrast, ConvNeXt and DeiT show no significant difference ($p=0.389$), suggesting comparable decision behavior. These observations indicate that the three backbone models capture distinct yet partially overlapping feature representations of deepfake artifacts, reflected in their areas of disagreement. 
\begin{table*}[htbp]
    \centering
    \caption{Results of McNemar statistical test comparing pairwise classifier performance}
    \label{tab:Mc-Nemar Statistical Analysis}
    \resizebox{\textwidth}{!}{
    \begin{tabular}{l l c c c c c c c}
    \hline
     \multirow{2}{*}{Model 1} & \multirow{2}{*}{Model 2} & $n_{11}$ & $n_{10}$ & $n_{01}$ & $n_{00}$ & \multirow{2}{*}{$\chi^2$} & p-value & \multirow{2}{*}{Significant}\\
     & & (both correct) & (M1 correct) & (M2 correct) & (both wrong) & & ($\alpha=0.05$) \\
     \hline
      EfficientNet & ConvNeXt & 2576 & 213 & 173 & 110 &   3.94 & 0.047 & Yes\\ 
      EfficientNet & DeiT & 2572 & 217 & 159 & 124 & 8.641 & 0.003 & Yes\\
      ConvNeXt & DeiT & 2545 & 204 & 186 & 137 & 0.741 & 0.389 & No\\
      EfficientNet & CAE-Net & 2752 & 37 & 145 & 138 & 62.907 & 0.000 & Yes\\
      ConvNeXt & CAE-Net & 2720 & 29 & 177 & 146 & 104.898 & 0.000 & Yes\\
      DeiT & CAE-Net & 2701 & 30 & 196 & 145 & 120.465 & 0.000 & Yes\\
      \hline
    \end{tabular}}
\end{table*}
The ensemble model (CAE-Net), however, demonstrates a clear and statistically significant improvement over all three individual models, with $p$-values approaching zero even after Bonferroni correction. Specifically, the ensemble achieves 108, 148, and 166 additional correct classifications compared to EfficientNet, ConvNeXt, and DeiT, respectively, in the disagreement cases where differential performance matters most. Moreover, the relatively low number of disagreements between the ensemble and base models (182--226 cases) suggests that the ensemble rarely errs when base models succeed, but frequently corrects their mistakes. Overall, these results provide strong statistical evidence that the ensemble’s weighted fusion mechanism effectively leverages the complementary strengths of its constituent architectures. The ensemble not only achieves higher accuracy but also exhibits more consistent and reliable decision-making, confirming that its performance gains are statistically robust and unlikely due to chance.

\subsection{Comparison with the existing works}
There exist very few works on deepfake detection that perform well on the combination of such a diverse set of datasets. A comparative analysis among existing works with our proposed architecture is shown in Table~\ref{Comparison_table}. Note that the AUC score presented here is the mean score of these models on the corresponding datasets (we refer the reader to Figure 8 of~\cite{li2020celeb}). We want to emphasize that existing works often demonstrate good performance on only a few datasets, but the overall performance on multiple datasets is poor for most architectures.
\begin{table*}[htbp]
    \centering
    %\scriptsize
    \caption{Comparative analysis of average AUC score of different models}
    
    \label{Comparison_table}
    \resizebox{\textwidth}{!}{
    \begin{tabular}{c c c c}
        \hlineB{2}
        \textbf{Detection Method} & \textbf{Datasets} & \textbf{AUC Score} & \textbf{Std. Dev.}\\
        \hlineB{2}
        \multirow{2}{*}{Capsule~\cite{nguyen2019use}} & UADFV~\cite{li2018ictu}, DF-TIMIT~\cite{korshunov2018deepfakes}, FaceForensics++~\cite{rossler2019faceforensics++}, & \multirow{2}{*}{69.4\%} &\multirow{2}{*}{13.84\%}\\
         & DFD~\cite{Dufour_Gully_2019}, DFDC~\cite{dolhansky2020deepfake}, Celeb-DF~\cite{li2020celeb} & & \\
        \hline
        
        \multirow{2}{*}{Meso4~\cite{afchar2018mesonet}} & UADFV~\cite{li2018ictu}, DF-TIMIT~\cite{korshunov2018deepfakes}, FaceForensics++~\cite{rossler2019faceforensics++}, & \multirow{2}{*}{75.9\%} &\multirow{2}{*}{10.63\%}\\
         & DFD~\cite{Dufour_Gully_2019}, DFDC~\cite{dolhansky2020deepfake}, Celeb-DF~\cite{li2020celeb} & & \\
        \hline
        
        \multirow{2}{*}{FWA~\cite{li2018exposing}} & UADFV~\cite{li2018ictu}, DF-TIMIT~\cite{korshunov2018deepfakes}, FaceForensics++~\cite{rossler2019faceforensics++}, & \multirow{2}{*}{82.1\%} &\multirow{2}{*}{14.45\%}\\
         & DFD~\cite{Dufour_Gully_2019}, DFDC~\cite{dolhansky2020deepfake}, Celeb-DF~\cite{li2020celeb} & &\\
        \hline
        
        \multirow{2}{*}{Xception-C23~\cite{rossler2019faceforensics++}} & UADFV~\cite{li2018ictu}, DF-TIMIT~\cite{korshunov2018deepfakes}, FaceForensics++~\cite{rossler2019faceforensics++}, & \multirow{2}{*}{86.4\%} &\multirow{2}{*}{11.96\%}\\
        & DFD~\cite{Dufour_Gully_2019}, DFDC~\cite{dolhansky2020deepfake}, Celeb-DF~\cite{li2020celeb} & &\\
        \hline
        
        \multirow{2}{*}{DSP-FWA~\cite{li2020celeb}} & UADFV~\cite{li2018ictu}, DF-TIMIT~\cite{korshunov2018deepfakes}, FaceForensics++~\cite{rossler2019faceforensics++}, & \multirow{2}{*}{87.4\%} &\multirow{2}{*}{12.79\%}\\
         & DFD~\cite{Dufour_Gully_2019}, DFDC~\cite{dolhansky2020deepfake}, Celeb-DF~\cite{li2020celeb} & &\\
        \hline
        \multirow{2}{*}{Our Proposed Method} & Celeb-DF~\cite{li2020celeb}, DFDC~\cite{dolhansky2020deepfake}, DFD~\cite{Dufour_Gully_2019}, UADFV~\cite{li2018ictu}, & \multirow{2}{*}{\textbf{97.37\%}} &\multirow{2}{*}{-}\\
         & FaceShifter~\cite{li2019faceshifter}, DFDC Preview~\cite{dolhansky2019dee}, FaceForensics++~\cite{rossler2019faceforensics++} & &\\
        \hline
    \end{tabular}
    }
\end{table*}
We compared our proposed model with existing ensemble and frequency–spatial hybrid architectures on the DF-Wild Cup dataset to evaluate both classification performance and model complexity. As shown in Table~\ref{tab:comparison}, CAE-Net achieves superior performance, attaining 94.46\% accuracy with 117.4M parameters, outperforming previous methods, such as ResNeXt–Xception, DeepfakeStack, and CoAt-Net ensembles. Unlike prior approaches that primarily focus on either spatial or hybrid CNN–Transformer integration without frequency-domain enhancement, CAE-Net incorporates wavelet-based preprocessing to extract complementary frequency cues while maintaining architectural diversity among its sub-models. This balanced design allows the ensemble to learn richer representations of deepfake artifacts and to generalize more effectively across diverse manipulation techniques. Consequently, CAE-Net achieves a better trade-off between accuracy and computational cost, demonstrating the effectiveness of its ensemble logic compared to existing state-of-the-art models.

\begin{table}[htbp]
    \centering
    \caption{Comparative analysis of our work with prior ensembles and frequency-spatial hybrid networks}
    \label{tab:comparison}
    \resizebox{0.6\textwidth}{!}{
    \begin{tabular}{l c c}
    \hline
    Classifiers  & \# of params. & Accuracy\\
    \hline
    Freq-Net~\cite{tan2024frequency} & 1.85 M & 67.35\%\\
    FSBI~\cite{hasanaath2025fsbi} & 23.34 M & 82.00\%\\
    CoAt-Net Ensemble~\cite{omar2024ensemble} & 699.72 M & 84.22\%\\
    SFA-Net~\cite{liu2025deepfake} & 25.04 M & 90.07\%\\
    ResNeXt \& Xception Ensemble~\cite{gowda2022investigation} & 43.8 M & 91.21\%\\
    DeepfakeStack~\cite{rana2020deepfakestack} & 135.55 M & 92.12\%\\
    \textbf{Proposed CAE-Net} & 117.4 M & \textbf{94.46\%}\\
    \hline
    \end{tabular}}
\end{table}

\subsection{Robustness analysis against adversarial attacks}
Finally, we look into the effectiveness of our ensemble method through a simple adversarial perturbation attack, such as Fast Gradient Sign Method (FGSM). As mentioned before, FGSM can fool models by adding simple perturbation noise based on the gradient sign that intends to maximize the loss function of the detection model instead of minimizing it. We present the accuracy scores of the separate detection model as well as the ensemble models for different values of $\epsilon$ in Table~\ref{FGSM_table}.\\
\begin{table*}[t]
    \caption{Comparison of our proposed architectures under FGSM attacks}
    \label{FGSM_table}
    \centering
    \resizebox{\textwidth}{!}{
    \begin{tabular}{l c c c c | c c c c c}
    \hline
    \multirow{2}{*}{\textbf{Model}} & \multicolumn{4}{c}{\textbf{Accuracy (before adversarial training)}} & \multicolumn{5}{c}{\textbf{Accuracy (after adversarial training)}} \\
    \cline{2-5}
    \cline{6-10}
    & $\epsilon = 0.005$ & $\epsilon = 0.01$ & $\epsilon = 0.03$ & $\epsilon = 0.05$ & $\epsilon = 0.00$ & $\epsilon = 0.005$ & $\epsilon = 0.01$ & $\epsilon = 0.03$ & $\epsilon = 0.05$ \\
    \hline
     EfficientNetB0 & 42.77\% & 38.54\% & 46.48\% & 53.94\% & 93.07\% & 56.90\% & 50.39\% & 45.21\% & 50.26\%\\
     DeiT & 49.61\%  & 42.76\% & 21.94\% & 16.73\% & 89.29\% & 65.10\% & 37.14\% & 5.57\% & 3.74\%\\
     ConvNeXt with wavelet & 16.08\% & 6.45\% & 6.51\% & 15.40\% & 92.25\% & 98.40\% & 96.45\% & 63.28\% & 51.63\%\\
     \hline
     Weighted Ensem.(CAE-Net) & 20.64\% & 12.04\% & 13.48\% & 16.05\% & \textbf{94.43\%} & \textbf{87.27\%} & \textbf{69.92\%} & 14.84\% & 20.25\%\\
     Majority Voting (CAE-Net) & \textbf{34.86\%} & \textbf{25.52\%} & \textbf{20.41\%} & \textbf{19.76\%} & 93.85\% & 82.58\% & 64.29\% & \textbf{25.03\%} & \textbf{35.52\%}\\
    \hline
    \end{tabular}
    }
\end{table*}
The results show an increased robustness of the EfficientNet architecture, which depicts increased accuracy with increasing perturbation compared to the other two architectures. DeiT shows a decrease in accuracy with rising perturbation, and ConvNeXt accuracy drops significantly even for very low perturbations added to the image, which is likely caused by the modification of frequency domain artifacts by even small attacks, which disrupts its feature extraction. Though a weighted ensemble model is appropriate for achieving high accuracy on the generalized dataset, we argue that it is highly vulnerable to adversarial attacks. If any model performs poorly on adversarial data, then the performance of the entire ensemble drops significantly. However, in the majority voting ensemble, the performance does not reduce as much, since it requires the majority of models to perform poorly, implying moderate robustness against adversarial attacks for applications, such as deepfake detection. 

To improve the robustness of our models against adversarial attacks, we integrated adversarial training using pre-trained weights optimized for clean data. This involved training with adversarial samples generated using a small perturbation ($\epsilon = 0.005$) over six epochs, employing a reduced learning rate of 1e-5 and a batch size of 64. We adopted a mixed batch strategy, comprising 32 perturbed images and 32 unperturbed images per batch, to balance exposure to both conditions~\cite{goodfellow2014explaining}. Notably, this approach yielded significant enhancements, with models achieving improved accuracy on the clean validation set and demonstrating superior resilience to smaller FGSM perturbations ($\epsilon = 0.005, 0.01$) without adversarial training. The weighted ensemble and majority voting also show enhanced resilience to the FGSM attacks, with accuracy shifting from 20.64\% without adversarial training to 87.27\% (see Table~\ref{FGSM_table}) after adversarial training for $\epsilon = 0.005$. The increase in overall accuracy is maintained for higher perturbations as well. Again, majority voting outperforms weighted ensemble in the case of higher perturbations, which is expected. Further training with incrementally larger perturbations and with more sophisticated adversarial images (such as, projected gradient descent) may enhance performance against higher perturbation levels, a direction we intend to explore in future work.

\section{Discussion}
\label{sec: Discussion}

This study presents CAE-Net, an ensemble architecture combining convolutional and attention-based models with spatial and frequency domain features for generalized deepfake detection. In this section, we discuss our design decisions, analyze the deployment implications of our results, acknowledge the limitations of our work, and outline future research directions.

The first major challenge of our study was addressing the 5:1 class imbalance in our dataset. Initially, we attempted conventional techniques like random oversampling and weighted loss functions. However, these techniques presented critical limitations, as stated in Section~\ref{class balancing}. Our method ensures balanced 1:1 exposure at each of the five training stages, with each stage using all 42,690 real images paired with a disjoint subset of 43,894 fake images. Through this method, the authenticity of real images is preserved, and knowledge accumulation is continued via warm starting the next stages. Recent estimates suggest AI-generated content now exceeds human-created content in volume online~\cite{GraphiteAIContent2023}. As deepfake generation becomes more accessible, detection datasets will increasingly exhibit class imbalance, with synthetic samples vastly outnumbering authentic ones. Our disjoint multistage training method provides a scalable solution for such scenarios without requiring proportional increases in authentic training data.

We position CAE-Net for deployment scenarios where accuracy is very important and computational resources are available, such as centralized content moderation systems or forensic laboratories. For resource-constrained environments, such as edge devices, the individual EfficientNet model provides a viable alternative with only 3.96 M parameters and 386 M FLOPs, as shown in Table~\ref{tab: Ensemble ablation}. For scenarios where resource constraints are more relaxed, the ensemble of EfficientNet and ConvNeXt is also a good option with a 1.5s inference time and 31.77 M parameters with a high accuracy of 93.14\%.

Note that the ensemble of EfficientNet and ConvNeXt provides detection accuracy comparable to CAE-Net with less than one-fourth the parameters and half the inference time. However, we emphasize here the 1.32\% accuracy improvement of CAE-Net is equivalent to correctly identifying 13,200 additional deepfakes per million processed images in content moderation systems. Given the serious consequences of deepfake spread, including financial fraud, electoral interference, and non-consensual imagery, the requirement for high accuracy justifies our ensemble approach's computational expense in critical applications.

Our evaluation under FGSM attacks revealed that CAE-Net's accuracy drops significantly under perturbations for the weighted ensemble, while majority voting performs slightly better under the same attack. This vulnerability stems from our probability-level fusion mechanism: when adversarial perturbations cause even one model's output probabilities to shift significantly, the weighted sum propagates this error proportionally. However, the binary decision mechanism of majority voting scheme  provides a bounded impact where a single compromised model cannot override the other two. This represents a fundamental trade-off between clean-data optimization (weighted ensemble) and adversarial resilience (majority voting). However, upon adversarial training, the models show better robustness against such attacks, outlining the importance of such enhancement methods for deepfake classifiers. For deployments in adversarially exposed environments such as public-facing platforms, we recommend majority voting. For accuracy-critical applications with lower adversarial threat, the weighted ensemble remains preferable.

CAE-Net employs a probability-level fusion rather than intermediate feature-level fusion. Each model (EfficientNet, ConvNeXt, DeiT) processes inputs independently through its complete architecture, maintaining native feature representations without cross-model feature map alignment or shared layers. This design choice offers key advantages: architectural independence allows each model to be trained and optimized separately, computational efficiency is obtained through parallel inference, and the modularity of late fusion enables straightforward addition or removal of ensemble members. Fusion occurs only at the final layer through weighted averaging of class probabilities, making the ensemble mechanism interpretable and the contribution of each model directly quantifiable.\\

\textbf{Limitations and Future Directions. }
Several limitations of our work require discussion. First, the dataset provided by the IEEE SPCup 2025 organizers had a fixed validation set. While defining the weights of the ensemble, we could not implement cross-validation due to this fixed split, which limits the robustness of our weight selection. Second, hyperparameter optimization was performed manually due to resource constraints rather than through systematic grid search or automated methods. Third, while testing the benchmark model performances described in Table~\ref{tab:CNN experimentations},~\ref{tab:transformer experimentations},~\ref{tab:frequency domain experimentations}, we used a fixed random seed of 42 for reproducibility and fair comparison. While this ensures consistent results for that specific initialization, it does not guarantee that our observed improvements generalize across different random initializations.

The Grad-CAM visualizations indicate that the models sometimes concentrate on irrelevant areas, such as the background or non-facial features, instead of the artifact-prone facial regions essential for accurate classification. This misdirected focus contributes to misclassifications. To address this issue, incorporation of attention regularization techniques can work to guide the models toward critical facial areas, prioritizing zones with significant artifacts. Additionally, our evaluation does not assess performance on compressed or degraded images, which are common in real-world social media contexts. The robustness of frequency-domain features to such compression artifacts requires further investigation. Our ablation studies focused on architectural combinations but did not explore alternative fusion strategies beyond weighted averaging and majority voting, such as learned meta-classifiers, attention-based fusion, or stacking. 
Future research directions include developing efficient model compression techniques, such as knowledge distillation, quantization, and pruning, to enable deployment on resource-constrained devices while maintaining high accuracy, and testing generalization to emerging deepfake generation methods, including diffusion models and neural radiance fields (NeRFs)~\cite{mildenhall2021nerf}.

\section{Conclusion}
\label{sec:conc}
In this work, we proposed an ensemble architecture for deepfake detection that captured the intricate properties of the input images to successfully distinguish between real and fake images and produced excellent results. We emphasize that detecting deepfake images that are generated with a range of generative AI approaches is a challenging task. To that end, we have addressed a significant class imbalance through disjoint training set-based multistage training and proposed a novel weighted ensemble approach that takes advantage of the strengths of an EfficientNet model, a Data Efficient Image Transformer model, and a ConvNext model for the accurate classification of deepfake images. The EfficientNet-B0 model enabled extracting local features from an image, while the DeiT model specialized in capturing global features. Additionally, ConvNeXt, combined with the wavelet transform, effectively identified frequency-dependent artifacts. Statistical analysis shows that the CAE-Net significantly outperforms the constituent backbone models. Through t-SNE plots, we demonstrated that our model is capable of separating real and fake images into separable clusters. Additionally, we included Grad-CAM visualizations to map the focus regions of images to illustrate how our models analyze the images and which portions contribute to distinguishing between real and fake images. Besides, we further analyzed the effectiveness of the proposed models by testing them against FGSM attacks and found the optimum ensemble mechanism to be majority voting compared to the weighted ensemble with a little trade-off in accuracy. An interesting future direction could be the incorporation of lightweight architectures, which would reduce computational overhead and increase the model's adaptability to newer and more sophisticated deepfake generation methods. \\

\small
\bibliographystyle{ieeetr}
\bibliography{references}
%%--------------------------------------- End references 

%
\end{document}